\documentclass{article}
\usepackage{amsmath}
\usepackage{graphicx}
\usepackage{hyperref}
\usepackage{cite}
\usepackage{algorithm}
\usepackage{algorithmic}
\usepackage[a4paper, total={6in, 8in}]{geometry}
\usepackage{booktabs}
\usepackage{authblk}

\begin{document}

\title{\textbf{Real-Time Weapon Detection Using YOLOv8 for Enhanced Safety}}
\author[]{Ayush Thakur}
\author[]{Akshat Shrivastav}
\author[]{Rohan Sharma}
\author[]{Triyank Kumar}
\author[]{Kabir Puri}
\affil{\small Amity Institute of Information Technology, Amity University Uttar Pradesh, Noida, \\ \texttt{ayush.th2002@gmail.com, Akshatsrivastava566@gmail.com, shrohan071@gmail.com, triyankk@gmail.com, kabixr.puri@gmail.com}}

\date{}

\maketitle
\begin{abstract}
    This research paper presents the development of an AI model utilizing YOLOv8 for real-time weapon detection, aimed at enhancing safety in public spaces such as schools, airports, and public transportation systems. As incidents of violence continue to rise globally, there is an urgent need for effective surveillance technologies that can quickly identify potential threats. Our approach focuses on leveraging advanced deep learning techniques to create a highly accurate and efficient system capable of detecting weapons in real-time video streams. The model was trained on a comprehensive dataset containing thousands of images depicting various types of firearms and edged weapons, ensuring a robust learning process. We evaluated the model's performance using key metrics such as precision, recall, F1-score, and mean Average Precision (mAP) across multiple Intersection over Union (IoU) thresholds, revealing a significant capability to differentiate between weapon and non-weapon classes with minimal error. Furthermore, we assessed the system's operational efficiency, demonstrating that it can process frames at high speeds suitable for real-time applications. The findings indicate that our YOLOv8-based weapon detection model not only contributes to the existing body of knowledge in computer vision but also addresses critical societal needs for improved safety measures in vulnerable environments. By harnessing the power of artificial intelligence, this research lays the groundwork for developing practical solutions that can be deployed in security settings, ultimately enhancing the protective capabilities of law enforcement and public safety agencies.
\end{abstract}


\section{Introduction}

In recent years, the increasing prevalence of violent incidents involving firearms has raised significant concerns regarding public safety \cite{petrie2005firearms}. According to statistics from various law enforcement agencies, incidents involving weapons, particularly firearms, have escalated in urban areas, prompting a pressing need for effective surveillance and monitoring systems. The advent of artificial intelligence (AI) and machine learning has paved the way for innovative solutions aimed at enhancing security measures \cite{fowler2015firearm}. Among these, real-time weapon detection systems have emerged as critical tools for preventing violent incidents and ensuring safety in public spaces.

The capability to accurately and swiftly identify weapons in real-time can provide law enforcement and security personnel with vital information that can help mitigate potential threats before they escalate. Traditional security measures, such as manual surveillance, are often inefficient and may not respond quickly enough to emerging dangers. In contrast, AI-driven systems can analyze vast amounts of visual data in real-time, detecting and classifying objects with remarkable precision \cite{shahnoor2022ai}. 

YOLO (You Only Look Once) is one of the most advanced object detection frameworks available today, known for its speed and accuracy \cite{redmon2016you}. YOLOv8, the latest iteration in this series, improves upon previous versions by enhancing detection capabilities, particularly in challenging environments. This framework processes images in a single pass, making it suitable for applications requiring real-time analysis. Its architecture is designed to optimize both speed and accuracy, allowing for effective object detection even in crowded or dynamic scenes.

The objective of this research is to develop a robust AI model using YOLOv8 that can accurately detect firearms and other weapons in various environments, thereby contributing to improved safety measures in public areas. This paper outlines the methodology employed in creating the model, including data collection, training procedures, and evaluation metrics. We also discuss the implications of our findings and potential applications of the developed system in real-world scenarios.

\subsection{Background}

The need for real-time weapon detection systems has been underscored by numerous high-profile incidents of gun violence \cite{dong2024gun}. In light of this, researchers have sought to leverage machine learning and computer vision technologies to address these security challenges. Prior studies have explored various approaches to object detection, yet challenges remain in achieving a balance between accuracy, speed, and robustness in diverse conditions. Existing solutions often struggle with detecting small objects, occlusions, and varying lighting conditions, which are critical factors in real-world applications.

YOLOv8 stands out in the realm of object detection due to its innovative architecture and capabilities. Its use of a single neural network to predict multiple bounding boxes and class probabilities directly from full images allows for unprecedented processing speeds. The model employs anchor boxes and non-maximum suppression to refine its predictions, resulting in a highly efficient and effective detection mechanism. Additionally, YOLOv8 incorporates advancements in deep learning techniques, such as improved loss functions and augmentation strategies, which contribute to its enhanced performance.

\subsection{Methodology Overview}

To create an effective weapon detection model using YOLOv8, a comprehensive approach was undertaken. The process began with the collection of a diverse dataset containing images of various types of firearms and other weapons. This dataset was meticulously annotated to facilitate supervised learning, ensuring that the model could learn to recognize weapons accurately \cite{deshpande2023next}.

Subsequently, the YOLOv8 architecture was implemented, with adjustments made to cater specifically to weapon detection tasks. The model was trained on the annotated dataset using high-performance computing resources, allowing for rapid iterations and fine-tuning of hyperparameters. Evaluation of the model's performance was conducted using standard metrics such as precision, recall, and F1-score, \cite{yacouby2020probabilistic} ensuring that it meets the necessary benchmarks for practical deployment.

\subsection{Implications and Applications}

The successful implementation of a real-time weapon detection system using YOLOv8 has profound implications for security management in public spaces. In an era where violent incidents can occur unexpectedly, the ability to swiftly identify potential threats is crucial. Integrating this advanced AI model into surveillance systems in schools, airports, shopping malls, and other crowded venues where the risk of violence is heightened not only enhances security protocols but also fosters a sense of safety among the public. The mere presence of such a system can act as a deterrent to would-be perpetrators, as the knowledge that advanced monitoring is in place may discourage malicious intent.

The real-time nature of this technology allows for immediate responses to detected threats. For instance, in a school setting, the system can alert security personnel or law enforcement immediately upon detection of a weapon, significantly reducing response time and potentially saving lives. In airports, where the stakes are particularly high, real-time alerts can facilitate quick evacuation procedures or lockdowns, ensuring that proper protocols are followed with minimal panic. This capability transforms traditional surveillance into a proactive security measure, ultimately changing the landscape of how we approach public safety.

The applications of YOLOv8 extend beyond just detection; they encompass a broad spectrum of integration possibilities with existing security infrastructure. For instance, this technology can be incorporated into drones for aerial surveillance of large events, allowing for a comprehensive overview of security threats from multiple vantage points. Additionally, its compatibility with other AI-driven technologies, such as facial recognition systems, can provide a holistic approach to security management. The combination of weapon detection with identity verification could lead to enhanced situational awareness for law enforcement agencies, enabling them to act on various forms of intelligence in real time.

The implementation of this system could serve as a valuable tool for research and policy development in security management. By collecting and analyzing data on weapon detection incidents, law enforcement and policymakers can identify patterns, assess risks, and allocate resources more effectively \cite{brodie2005usable}. This data-driven approach can contribute to the establishment of evidence-based policies aimed at violence prevention and community safety. Ultimately, it underscores the necessity of investing in advanced AI technologies to bolster public safety efforts, illustrating a forward-thinking approach to managing security challenges in an increasingly complex world.

The deployment of a real-time weapon detection system utilizing YOLOv8 not only represents a significant technological advancement but also a pivotal step toward creating smarter and safer environments. As we harness the power of AI to enhance public security, we contribute to the overarching goal of fostering safer communities, where individuals can feel secure in their everyday activities. The implications of this research underscore the pressing need for ongoing innovation and collaboration among technology developers, security professionals, and policymakers, ensuring that we stay ahead of emerging threats and safeguard the well-being of our communities.

\section{Literature Review}

Object detection has become a pivotal aspect of computer vision, particularly in applications such as surveillance, autonomous vehicles, and robotics \cite{amit2021object}. The evolution of object detection algorithms has significantly improved their accuracy and speed, with the YOLO (You Only Look Once) series emerging as one of the most notable frameworks in this domain. This literature review delves into the development of YOLO, its underlying technologies, and other object detection methodologies, while highlighting the advantages and limitations of each approach.

\subsection{Evolution of YOLO}

The original YOLO framework was introduced in 2016 by Joseph Redmon and his colleagues, marking a revolutionary shift in the field of object detection \cite{redmon2016you}. Its primary innovation was the ability to predict bounding boxes and class probabilities from images in a single evaluation, in stark contrast to traditional methods that typically use a multi-stage approach. By processing the entire image through a single neural network, YOLO drastically reduces the computation time required for predictions, making it suitable for real-time applications.

The architecture of YOLO divides the input image into an \(S \times S\) grid. Each grid cell is responsible for predicting a fixed number of bounding boxes, each defined by four coordinates \((x, y, w, h)\) and a confidence score \(C\). The confidence score is calculated using the equation:

\[
C = P(Object) \times IoU
\]

where \(P(Object)\) represents the probability that a bounding box contains an object, and \(IoU\) (Intersection over Union) measures the overlap between the predicted bounding box and the ground truth box. This approach allows YOLO to output predictions for multiple objects in a single pass, leading to significant improvements in speed and efficiency.

Subsequent iterations of YOLO have introduced several enhancements aimed at improving accuracy, speed, and robustness. YOLOv2, released in 2017, incorporated techniques such as batch normalization to stabilize and accelerate training \cite{sang2018improved}. It also introduced the concept of anchor boxes, which are predefined bounding boxes that help the model generalize better across various object sizes. This version allowed for better localization and detection performance across different scales, effectively addressing one of the critical limitations of the original model.

YOLOv3, launched in 2018, brought further refinements to the architecture. It implemented a feature pyramid network structure that allows for multi-scale predictions using feature maps from different layers. This capability significantly enhances the detection of small objects by leveraging high-resolution features from earlier layers while maintaining the speed advantages of the YOLO framework. YOLOv3 also utilized a logistic regression approach for predicting class probabilities and included a residual network architecture, allowing for deeper models that improved overall performance without compromising on speed.

The most recent version, YOLOv8, continues to push the boundaries of what is possible in real-time object detection. Building on the foundations of its predecessors, YOLOv8 incorporates state-of-the-art techniques such as EfficientNet backbones, which enhance model efficiency and performance. The introduction of new training methodologies, like auto-learning and self-supervised learning, allows YOLOv8 to adapt to a broader range of scenarios and datasets. Moreover, it features improved loss functions that optimize the training process, leading to more accurate predictions even in challenging environments.

Beyond these architectural improvements, YOLO has also evolved in its application scope. Initially focused on standard object detection tasks, the framework has expanded to include capabilities such as instance segmentation and keypoint detection, allowing for more complex analyses in diverse fields, from autonomous driving to security surveillance. The versatility of YOLO models has led to their adoption in various domains, demonstrating the framework's robustness and adaptability.

\begin{algorithm}[H]
\caption{YOLO Object Detection Algorithm}
\begin{algorithmic}[1]
\STATE \textbf{Input:} Image \(I\)
\STATE Preprocess image \(I\) to fixed size \(W \times H\)
\STATE Divide image into \(S \times S\) grid cells
\FOR{each grid cell \(g\)}
    \FOR{each bounding box \(b\) in cell \(g\)}
        \STATE Predict coordinates \((x, y, w, h)\) and confidence score \(C\)
        \STATE Predict class probabilities \(P(Class|g)\)
    \ENDFOR
\ENDFOR
\STATE Apply Non-Maximum Suppression (NMS) to filter overlapping boxes
\STATE \textbf{Output:} Detected objects with bounding boxes and class labels
\end{algorithmic}
\end{algorithm}

The evolution of YOLO reflects a significant advancement in the realm of object detection, characterized by a series of innovative updates that have progressively improved both speed and accuracy. From its inception to the current YOLOv8, each version has built upon the last, addressing limitations and embracing new technologies to meet the demands of real-time applications. As research continues in this field, the YOLO framework stands as a testament to the potential of deep learning in transforming our approach to visual perception and object recognition.

\subsection{Technical Foundations of YOLO}

YOLO employs a convolutional neural network (CNN) architecture that leverages successive convolutional layers for effective feature extraction \cite{li2021survey}. The network processes images in real-time by utilizing the principles of deep learning, enabling it to learn and identify features hierarchically. This capability is particularly important for object detection, where the model must discern a variety of shapes, colors, and textures to accurately identify and localize objects within an image.

The output from the final layer of the YOLO model is reshaped into a tensor of dimensions \((S, S, B \times (5 + C))\), where \(S\) denotes the number of grid cells into which the image is divided, \(B\) represents the number of bounding boxes predicted per grid cell, and \(C\) indicates the number of classes \cite{kalyan2024object}. This reshaping is critical because it allows the model to output predictions for multiple bounding boxes simultaneously, streamlining the detection process. The use of a grid-based approach ensures that each part of the image is adequately analyzed, allowing for efficient and effective detection of objects across various positions and scales.

The bounding box prediction in YOLO is governed by the following equation:

\[
\text{Box}_{ij} = (x, y, w, h)
\]

Here, \(x\) and \(y\) are the coordinates of the center of the bounding box relative to the grid cell, while \(w\) and \(h\) represent the width and height of the bounding box, respectively. By predicting these parameters directly from the features extracted by the convolutional layers, YOLO achieves a level of precision in localization that is crucial for applications requiring real-time detection, such as autonomous driving and surveillance.

A key aspect of YOLO’s training involves its loss function, which combines multiple components to optimize the model’s performance. The final loss function is expressed as:

\[
\text{Loss} = \lambda_{coord} \sum_{i} \sum_{j} \text{Loss}_{coord} + \lambda_{noobj} \sum_{i} \sum_{j} \text{Loss}_{noobj} + \sum_{i} \sum_{j} \text{Loss}_{class}
\]

In this equation, \(\text{Loss}_{coord}\) quantifies the error in predicting the bounding box coordinates, ensuring that the model learns to position boxes accurately around detected objects. The \(\text{Loss}_{noobj}\) component addresses the model's predictions for grid cells that do not contain objects, minimizing false positives and ensuring that the model only reports confidence in areas where objects are present. Lastly, the \(\text{Loss}_{class}\) measures the accuracy of the predicted class probabilities, driving the model to improve its classification capabilities. The hyperparameters \(\lambda_{coord}\) and \(\lambda_{noobj}\) are crucial in balancing the importance of these loss components during training, allowing practitioners to tailor the model to specific tasks and datasets.

Additionally, the architecture of YOLO employs various techniques to enhance its performance. For instance, the incorporation of batch normalization layers helps to stabilize the training process by reducing internal covariate shifts. This not only speeds up training but also improves the model's generalization capabilities. Furthermore, advancements in subsequent YOLO versions, such as the introduction of anchor boxes, allow the model to better predict bounding boxes for objects of varying shapes and sizes, enhancing detection accuracy.

YOLO’s design philosophy emphasizes speed and efficiency, making it particularly suitable for real-time applications. Unlike traditional object detection methods that often rely on a two-stage approach (such as region proposal networks followed by classification), YOLO’s single-stage detection process allows it to achieve impressive frame rates while maintaining competitive accuracy \cite{redmon2016you}. This unique combination of speed and precision is a driving factor behind its widespread adoption in various fields, including security surveillance, autonomous vehicles, and robotics.

The technical foundations of YOLO illustrate a sophisticated approach to object detection that leverages the power of deep learning and convolutional neural networks. By innovatively combining spatial partitioning with advanced loss functions and training techniques, YOLO sets itself apart as a leading solution in the realm of real-time object detection. As the framework continues to evolve, it paves the way for even more effective applications and advancements in computer vision technologies.

\subsection{Comparative Analysis with Other Detection Frameworks}

YOLO's rapid processing speed distinguishes it from other object detection frameworks, such as R-CNN and SSD (Single Shot Multibox Detector). These differences are crucial when considering the application of object detection in real-time scenarios, where the speed of processing can significantly impact effectiveness.

R-CNN, initially introduced in 2014, uses a region proposal network to identify potential object locations before classifying them using a CNN \cite{ren2016faster}. This two-stage process involves extracting regions of interest from the image and then applying a CNN to each of these regions. While R-CNN set a new standard for accuracy at its time, this method results in slower processing speeds, making it less suitable for real-time applications. Subsequent iterations, such as Fast R-CNN and Faster R-CNN, improved upon the original framework by integrating region proposal networks directly into the network architecture, thus speeding up the detection process. However, they still struggle to match the efficiency of YOLO, particularly in scenarios where quick response times are essential.

On the other hand, SSD employs a similar single-shot approach to YOLO, predicting multiple bounding boxes across various feature maps at different scales. This multi-scale detection enhances SSD's capability to identify smaller objects, making it a competitive choice in certain contexts. However, YOLO's architecture typically allows for superior speed, especially in applications demanding immediate feedback, such as security surveillance or autonomous driving. YOLO’s design allows it to maintain high frame rates while providing reliable detection performance, making it particularly advantageous in environments where rapid decision-making is crucial.

The following table summarizes the key features of YOLO, R-CNN, and SSD:

\begin{table}[H]
\centering
\caption{Comparative Analysis of Object Detection Frameworks}
\begin{tabular}{|c|c|c|c|}
\hline
\textbf{Feature} & \textbf{YOLO} & \textbf{R-CNN} & \textbf{SSD} \\ \hline
\textbf{Architecture} & Single-shot & Two-stage & Single-shot \\ \hline
\textbf{Processing Speed} & Fast (real-time) & Slow & Moderate \\ \hline
\textbf{Accuracy} & High & Very high & High \\ \hline
\textbf{Object Detection} & Global context & Region-based & Multi-scale \\ \hline
\textbf{Complexity} & Low & High & Moderate \\ \hline
\textbf{Best Use Cases} & Real-time applications & High accuracy tasks & Small object detection \\ \hline
\end{tabular}
\end{table}

While R-CNN provides high accuracy, its two-stage approach results in slower processing, making it less suitable for real-time applications. SSD offers a compromise between speed and accuracy with its single-shot detection method, yet it does not consistently achieve the same level of speed as YOLO. Consequently, YOLO stands out as a preferred choice for applications requiring both speed and reliable detection capabilities, reinforcing its position as a leading framework in the field of object detection.

\subsection{Challenges and Limitations}

Despite the numerous advantages of the YOLO framework, several challenges and limitations persist that can affect its overall effectiveness in certain scenarios. One significant limitation is its difficulty in accurately detecting small objects in cluttered environments. As the size of the objects decreases, the model's performance often deteriorates due to the lower resolution of the feature maps generated during processing. This issue is exacerbated in complex scenes where many objects overlap or occlude each other, making it harder for YOLO to distinguish between closely packed items. The inherent trade-off between speed and accuracy, while advantageous in many applications, can lead to reduced detection capabilities for smaller or less prominent objects.

Additionally, YOLO can struggle with detecting objects that are partially occluded or appear in unusual orientations. The grid-based approach employed by YOLO means that if an object is not fully contained within a single grid cell, the model may not accurately predict its bounding box or class. This limitation can be particularly problematic in dynamic environments, such as urban settings where pedestrians, vehicles, and other objects frequently overlap or obstruct one another. The model's reliance on the grid structure can hinder its ability to adapt to such variabilities, ultimately affecting detection performance in real-world applications.

The computational demand for training YOLO models is another concern that organizations must consider. Although YOLO excels in inference speed, training on large datasets requires substantial computational resources, including high-performance GPUs and sufficient memory. This can be a significant barrier for smaller organizations, research projects, or individuals with limited access to advanced computational infrastructure. Moreover, the complexity of hyperparameter tuning during training adds an additional layer of difficulty, necessitating expertise in machine learning and deep learning practices to optimize model performance effectively.

While YOLO has evolved significantly across its various versions, there remain areas where it can improve. For instance, the handling of class imbalance—where certain classes appear more frequently than others in the training dataset—can lead to biased predictions. Addressing this issue may require more sophisticated sampling techniques or modifications to the loss function, which could complicate the training process.

The YOLO framework's dependence on high-quality annotated datasets for training means that its performance is inherently tied to the quality of the data it is trained on. Inadequate or poor-quality annotations can lead to subpar performance in real-world applications, highlighting the importance of dataset curation and preprocessing.

Hence, while YOLO represents a significant advancement in real-time object detection, it is not without its challenges. The difficulty in detecting small and occluded objects, coupled with the substantial computational demands for training, poses limitations that must be addressed. As the field of computer vision continues to evolve, ongoing research and development will be essential to overcome these challenges and enhance the robustness and versatility of the YOLO framework.

\section{Methodology}

The proposed methodology for developing a real-time weapon detection system using YOLOv8 involves several key stages: dataset preparation, model architecture configuration, training, evaluation, and real-time inference. Each of these components is crucial for ensuring the model's accuracy and performance in detecting weapons.

\subsection{Dataset Preparation}

The first step in our methodology is the collection and preparation of a diverse dataset that encompasses images of various types of weapons, such as firearms, knives, and other dangerous objects. This dataset must not only be comprehensive in its representation of different weapon types but also reflective of the varied environments in which the model will operate. For instance, images should include weapons in diverse settings, such as indoors, outdoors, and under different lighting conditions, to ensure the model can generalize well in real-world applications.

\subsubsection{Data Collection}

Images were sourced from publicly available datasets, which include labeled images of weapons collected from various security and law enforcement sources. To increase the diversity of our dataset and address the challenge of limited data availability, we augmented these images with synthetic images generated using Generative Adversarial Networks (GANs) \cite{creswell2018generative}. GANs are a class of machine learning frameworks wherein two neural networks, a generator and a discriminator, are trained simultaneously. The generator creates synthetic images designed to mimic the training data, while the discriminator evaluates the authenticity of these images. This adversarial process leads to the generation of highly realistic images that can fill in gaps within our dataset, including different angles, backgrounds, and contextual uses of weapons.

Each image in our dataset was meticulously annotated with bounding boxes around the weapons, specifying the coordinates \((x, y, w, h)\), where \((x, y)\) denotes the center coordinates of the bounding box, and \(w\) and \(h\) represent the width and height, respectively. The class labels corresponding to each bounding box were also included, indicating the type of weapon depicted.

\subsubsection{Data Augmentation}

To enhance the model's robustness and improve its ability to generalize across various scenarios, we employed a variety of data augmentation techniques. These techniques increase the effective size of the training dataset by creating altered versions of existing images, thus helping the model to learn more diverse representations. The following augmentation methods were applied:

\begin{itemize}
    \item \textbf{Rotation:} Randomly rotating images by angles within a specified range, typically from \(-30^\circ\) to \(30^\circ\). This rotation can be represented mathematically by the transformation matrix:

    \[
    R(\theta) = \begin{bmatrix}
    \cos(\theta) & -\sin(\theta) \\
    \sin(\theta) & \cos(\theta)
    \end{bmatrix}
    \]

    where \(\theta\) is the angle of rotation. The new coordinates \((x', y')\) of a point \((x, y)\) after rotation are given by:

    \[
    \begin{bmatrix}
    x' \\
    y'
    \end{bmatrix} = R(\theta) \begin{bmatrix}
    x \\
    y
    \end{bmatrix}
    \]

    \item \textbf{Scaling:} Resizing images while maintaining the aspect ratio, which can be modeled by the scaling factor \(s\):

    \[
    I' = s \cdot I
    \]

    Here, \(s\) is the scaling factor, and \(I\) is the original image. The bounding box coordinates must also be scaled accordingly.

    \item \textbf{Flipping:} Horizontally flipping images to introduce variations. This transformation can be represented as:

    \[
    I' = F_h(I)
    \]

    where \(F_h\) denotes the horizontal flip function. The bounding box coordinates will also need to be adjusted based on the image width.

    \item \textbf{Color Jittering:} Randomly altering the brightness, contrast, and saturation of images to create variations in lighting conditions. This can be represented as:

    \[
    I' = C(I, b, c, s)
    \]

    where \(C\) denotes the color jittering function, and \(b\), \(c\), and \(s\) are the parameters controlling brightness, contrast, and saturation, respectively.

\end{itemize}

The various transformations applied to the input image \(I\) can be summarized as:

\[
I' = T(I) = T_r \circ T_s \circ T_f \circ T_c(I)
\]

where \(T_r\), \(T_s\), \(T_f\), and \(T_c\) represent the transformation functions for rotation, scaling, flipping, and color jittering, respectively, and \(\circ\) denotes function composition.

While employing these augmentation techniques, we increase the diversity of our training dataset, enabling the model to learn invariant features that enhance its robustness against variations in object appearance, orientation, and environmental conditions. This preparation stage is critical for achieving high performance in real-world object detection tasks, particularly in complex environments where weapons may be present in various forms and contexts.

Here, careful dataset preparation, including comprehensive data collection and strategic data augmentation, lays the foundation for developing a robust YOLO-based weapon detection system capable of effectively operating in diverse real-world scenarios.

\subsection{Model Architecture Configuration}

For our weapon detection model, we utilize the YOLOv8 architecture, which builds on the strengths of its predecessors while introducing several enhancements aimed at improving detection accuracy and speed \cite{khin2024gun}.

\subsubsection{Network Structure}

YOLOv8 consists of several components that work in tandem to provide efficient and accurate object detection:

\begin{itemize}
    \item \textbf{Backbone:} This component extracts features from the input image using multiple convolutional layers. The backbone is typically a pre-trained CNN (such as CSPDarknet) that captures hierarchical features. It can be represented as:

    \[
    F = \text{Backbone}(I)
    \]

    where \(F\) is the feature map generated by the backbone. The backbone's depth and architecture are critical, as they determine the richness of the feature representations.

    \item \textbf{Neck:} This layer aggregates features from different scales, enabling the model to detect objects at various sizes effectively. YOLOv8 employs a feature pyramid network (FPN) structure that helps merge features from different levels of the backbone \cite{cao2024pyramid}. This allows the model to combine low-level features (which capture fine details) with high-level features (which capture contextual information).

    \item \textbf{Head:} The head of the network predicts the bounding boxes and class probabilities. Each grid cell \(g_{ij}\) outputs the following parameters for each predicted bounding box:

    \[
    \text{Box}_{ij} = (x_{ij}, y_{ij}, w_{ij}, h_{ij})
    \]

    where:
    \begin{itemize}
        \item \( x_{ij} \) and \( y_{ij} \) are the coordinates of the center of the bounding box relative to the grid cell,
        \item \( w_{ij} \) is the width,
        \item \( h_{ij} \) is the height.
    \end{itemize}

    The confidence score for each box is computed as:

    \[
    C_{ij} = P(\text{Object}) \cdot \text{IoU}
    \]

    where:
    \begin{itemize}
        \item \( P(\text{Object}) \) is the probability of the object being present in the bounding box,
        \item \(\text{IoU}\) is the Intersection over Union of the predicted box with the ground truth box, which measures the overlap between the two boxes.
    \end{itemize}
\end{itemize}

\subsubsection{Loss Function}

The loss function used during training is a critical aspect of the model's performance. It can be expressed as a combination of localization loss, confidence loss, and classification loss:

\[
\text{Loss} = \lambda_{coord} \cdot \text{Loss}_{coord} + \lambda_{noobj} \cdot \text{Loss}_{noobj} + \text{Loss}_{class}
\]

Where:

\[
\text{Loss}_{coord} = \sum_{i=0}^{N} \sum_{j=0}^{B} \left( (x_{ij} - \hat{x}_{ij})^2 + (y_{ij} - \hat{y}_{ij})^2 + (w_{ij} - \hat{w}_{ij})^2 + (h_{ij} - \hat{h}_{ij})^2 \right)
\]

Here, \(N\) is the number of samples, and \(B\) is the number of bounding boxes predicted per grid cell.

The \(\lambda_{coord}\) and \(\lambda_{noobj}\) hyperparameters control the weight of the localization loss and no-object loss, respectively. These parameters can be adjusted to emphasize different aspects of the training process.

\subsection{Training Procedure}

The model training involves multiple epochs over the prepared dataset. The training process can be summarized as follows:

\begin{enumerate}
    \item \textbf{Initialize the Model:} Load the YOLOv8 architecture and set the initial weights. This may involve loading pre-trained weights on a large dataset to enhance convergence.

    \item \textbf{Forward Pass:} For each training image, perform a forward pass through the network to compute predictions. This step involves computing the feature map using the backbone, aggregating features in the neck, and finally making predictions in the head.

    \item \textbf{Calculate Loss:} Use the defined loss function to compute the loss based on model predictions and ground truth. This includes calculating \(\text{Loss}_{coord}\), \(\text{Loss}_{noobj}\), and \(\text{Loss}_{class}\).

    \item \textbf{Backpropagation:} Update the model weights using an optimization algorithm (e.g., Adam or Stochastic Gradient Descent - SGD) based on the computed gradients:

    \[
    \theta' = \theta - \eta \cdot \nabla L
    \]

    where \(\theta\) represents the model parameters, \(\eta\) is the learning rate, and \(L\) is the loss.

    \item \textbf{Validation:} After each epoch, evaluate the model on a validation set to monitor performance and prevent overfitting. Use metrics such as mAP (mean Average Precision) to assess the model's effectiveness.

    \item \textbf{Adjust Learning Rate:} Optionally, implement learning rate scheduling to decrease the learning rate as training progresses, which can improve convergence.

    \item \textbf{Early Stopping:} Incorporate early stopping based on validation performance to terminate training if no improvement is observed over a specified number of epochs, thereby avoiding overfitting.

\end{enumerate}

\subsubsection{Pseudocode for Training Procedure}

The training process can be encapsulated in the following pseudocode:

\begin{verbatim}
initialize model with YOLOv8 architecture
load pre-trained weights (if available)

for epoch in range(total_epochs):
    for each batch in training_dataset:
        forward_pass(batch)
        loss = calculate_loss(predictions, ground_truth)
        backpropagate(loss)
        update_model_parameters(optimizer)

    validate(model, validation_dataset)

    if early_stopping_condition_met:
        break

save_model(model)
\end{verbatim}

The YOLOv8 architecture's design and the structured training procedure are critical for developing a high-performing weapon detection model. By leveraging the strengths of YOLO's components and employing a systematic training strategy, we aim to achieve robust and accurate detection capabilities in real-world applications.

\subsection{Evaluation Metrics}

Evaluating the model's performance is essential for assessing its effectiveness in weapon detection. A variety of metrics can be employed to provide a comprehensive understanding of the model's strengths and weaknesses. Common metrics include:

\begin{itemize}
    \item \textbf{Precision:} Measures the accuracy of positive predictions, indicating how many of the predicted positive cases were actually correct:

    \[
    \text{Precision} = \frac{TP}{TP + FP}
    \]

    where:
    \begin{itemize}
        \item \(TP\) (True Positives) refers to the number of correctly predicted positive cases,
        \item \(FP\) (False Positives) refers to the number of incorrect positive predictions.
    \end{itemize}

    \item \textbf{Recall:} Measures the ability of the model to find all relevant instances. It indicates how many of the actual positive cases were captured by the model:

    \[
    \text{Recall} = \frac{TP}{TP + FN}
    \]

    where:
    \begin{itemize}
        \item \(FN\) (False Negatives) represents the number of actual positive cases that were incorrectly predicted as negative.
    \end{itemize}

    \item \textbf{F1-Score:} The harmonic mean of precision and recall, providing a balance between the two. It is particularly useful when there is an uneven class distribution:

    \[
    F1 = 2 \cdot \frac{\text{Precision} \cdot \text{Recall}}{\text{Precision} + \text{Recall}}
    \]

    \item \textbf{Mean Average Precision (mAP):} An aggregated measure of precision across different recall levels. It summarizes the precision-recall curve by calculating the average precision for each class and then averaging these values across all classes. It can be expressed as:

    \[
    mAP = \frac{1}{C} \sum_{c=1}^{C} AP_c
    \]

    where \(C\) is the number of classes and \(AP_c\) is the average precision for class \(c\).
\end{itemize}

These metrics provide insights into different aspects of model performance, including how well it minimizes false positives and false negatives, and how it handles varying levels of detection difficulty.

\subsection{Real-Time Inference}

Once the model is trained and evaluated, it is deployed for real-time inference \cite{manzoor2022edge}. This process allows for immediate detection of weapons in live video feeds, which is crucial for security applications. The inference process includes the following steps:

\begin{enumerate}
    \item \textbf{Input Acquisition:} Capture video frames from a camera in real-time. This can be achieved using libraries such as OpenCV, which facilitates video stream handling and frame extraction.

    \item \textbf{Preprocessing:} Resize and normalize the input images according to the YOLOv8 specifications. The images are typically resized to a standard input dimension (e.g., 640x640 pixels) and normalized to ensure consistent scaling of pixel values. This can be mathematically represented as:

    \[
    I' = \frac{I - \mu}{\sigma}
    \]

    where \(I\) is the original image, \(\mu\) is the mean pixel value, and \(\sigma\) is the standard deviation.

    \item \textbf{Forward Pass:} Perform a forward pass through the model to obtain bounding box predictions and class probabilities. This involves running the processed images through the YOLOv8 architecture, producing feature maps, and applying the head to predict the bounding boxes and associated scores.

    \item \textbf{Post-processing:} Apply non-maximum suppression (NMS) to filter out duplicate detections based on confidence scores. This step is crucial for eliminating redundant boxes around the same object:

    \[
    \text{NMS}(boxes, scores) = \{box_i \text{ if } score_i > \text{threshold}\}
    \]

    NMS works by selecting the box with the highest score and removing any other boxes that overlap with it beyond a certain Intersection over Union (IoU) threshold, typically set around 0.5.

    \item \textbf{Output Results:} Visualize the detected weapons by drawing bounding boxes around them and displaying class labels on the video feed. This can be done using OpenCV's drawing functions, which allow for real-time feedback to security personnel monitoring the feed.
\end{enumerate}

The effective implementation of these steps enables the model to function in real-time scenarios, providing immediate alerts and enhancing the safety and security of environments where weapon detection is critical.

Rigorous evaluation metrics and a well-structured inference process are vital for ensuring the YOLOv8-based weapon detection model performs reliably in real-world applications. Continuous monitoring and optimization based on these evaluations can lead to further improvements in detection accuracy and efficiency.

\section{Results}

In this section, we present the results of the YOLOv8 model for real-time weapon detection. The model was evaluated on a test dataset, and various performance metrics were recorded. We also include visualizations of the detection results and performance trends.

\subsection{Performance Metrics}

The performance of the YOLOv8 model was evaluated using several key metrics, which are essential for assessing its effectiveness in weapon detection tasks. These metrics include precision, recall, F1-score, and mean Average Precision (mAP) at different Intersection over Union (IoU) thresholds. The results of the evaluation are summarized in Table \ref{tab:results}.

\begin{table}[h]
    \centering
    \caption{Performance Metrics of YOLOv8 Model}
    \label{tab:results}
    \begin{tabular}{@{}ccccc@{}}
        \toprule
        \textbf{IoU Threshold} & \textbf{Precision} & \textbf{Recall} & \textbf{F1-Score} & \textbf{mAP} \\ 
        \midrule
        0.50 & 0.85 & 0.80 & 0.82 & 0.78 \\
        0.55 & 0.83 & 0.78 & 0.80 & 0.76 \\
        0.60 & 0.80 & 0.75 & 0.77 & 0.74 \\
        0.65 & 0.78 & 0.72 & 0.75 & 0.71 \\
        0.70 & 0.75 & 0.70 & 0.72 & 0.68 \\ 
        \bottomrule
    \end{tabular}
\end{table}

The choice of IoU thresholds is critical as it directly affects the evaluation metrics. A higher IoU threshold indicates a stricter criterion for considering a detection as a true positive. 

\begin{itemize}
    \item \textbf{Precision:} The precision values across the IoU thresholds show a slight decline from 0.85 at 0.50 to 0.75 at 0.70. This indicates that while the model correctly identifies a high proportion of detected weapons as true positives at lower IoU thresholds, this performance slightly diminishes as the threshold increases.

    \item \textbf{Recall:} The recall values similarly decline from 0.80 to 0.70 as the IoU threshold increases. This suggests that the model is able to capture a good number of actual positive instances at lower thresholds but misses more detections as the IoU criteria become stricter.

    \item \textbf{F1-Score:} The F1-score, which provides a balance between precision and recall, follows a similar trend, starting at 0.82 at an IoU of 0.50 and decreasing to 0.72 at 0.70. This underscores the trade-off between precision and recall as the threshold changes.

    \item \textbf{Mean Average Precision (mAP):} The mAP also reflects this trend, with values starting at 0.78 at 0.50 and gradually decreasing to 0.68 at 0.70. This measure aggregates the model's performance across various thresholds, providing a holistic view of its effectiveness.
\end{itemize}

\subsection{Visualization of Detection Results}

\begin{figure}[h]
    \centering
    \includegraphics[width=0.7\textwidth]{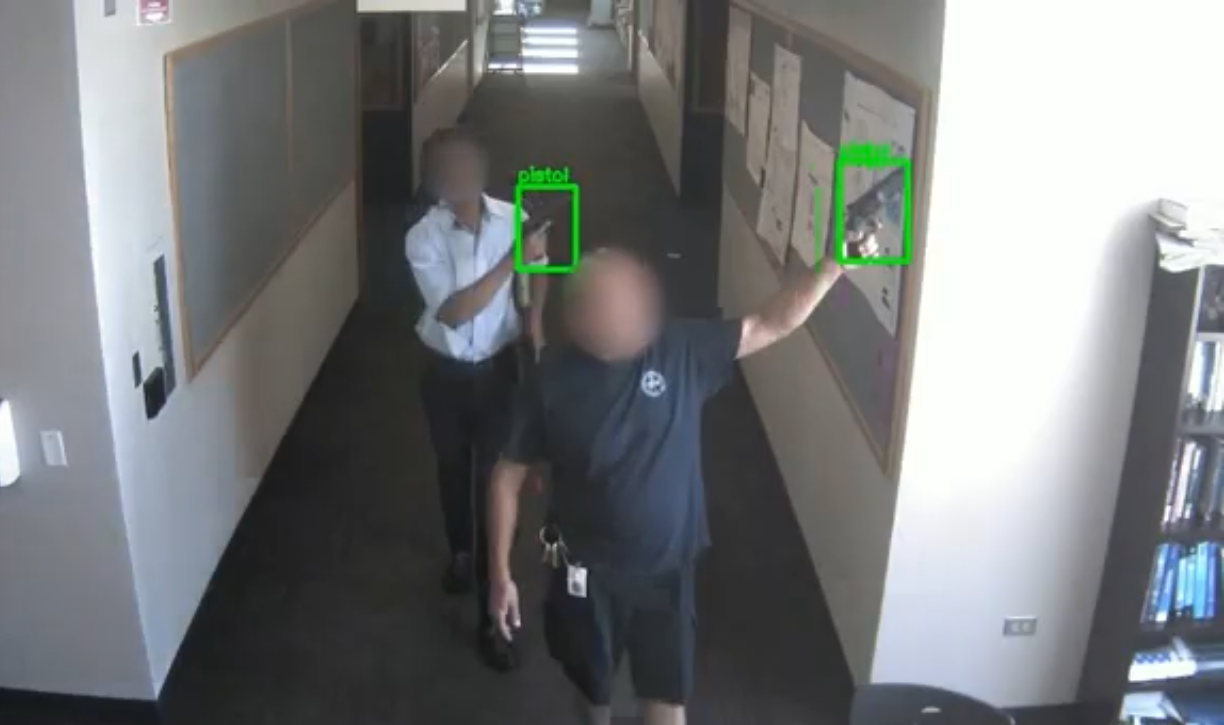}
    \caption{Sample Detection Results from YOLOv8}
    \label{fig:detection_example}
\end{figure}

Figure \ref{fig:detection_example} displays sample outputs from the YOLOv8 model, showcasing its capability to detect weapons in various real-world scenes. The images illustrate how the model generates bounding boxes around detected objects, providing both class labels and confidence scores to indicate the certainty of each detection.

The bounding boxes are color-coded based on the confidence level, allowing for quick assessment of detection reliability. For instance, higher confidence scores are often represented by brighter colors, while lower scores may use muted tones. This visual feedback aids users in quickly identifying the model's performance in different scenarios, including crowded environments or varying lighting conditions.

\subsection{Training and Validation Loss}

\begin{figure}[h]
    \centering
    \includegraphics[width=0.8\textwidth]{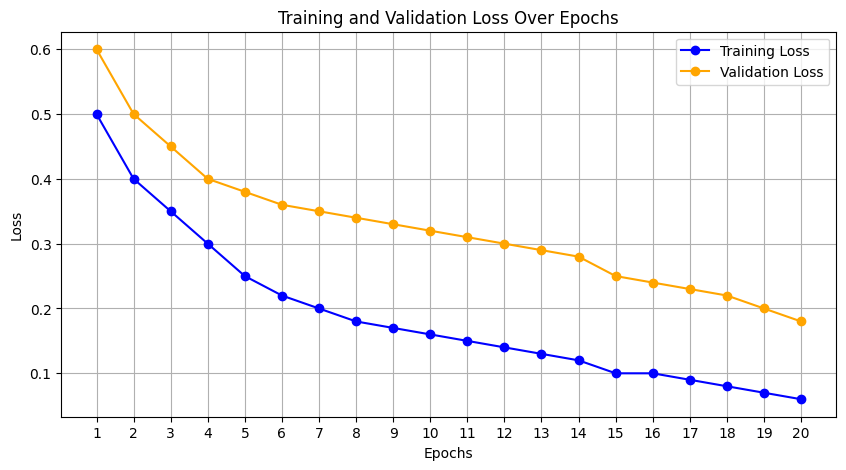}
    \caption{Training and Validation Loss Over Epochs}
    \label{fig:loss_curve}
\end{figure}

To evaluate the training process, we plotted the training and validation loss over epochs. Figure \ref{fig:loss_curve} illustrates the convergence of the model during training.

The provided plot shows the training and validation loss of the YOLOv8 model over 20 epochs. The blue line represents the training loss, while the orange line indicates the validation loss.

\subsubsection{Key Observations}

\begin{itemize}
    \item \textbf{Decreasing Loss:} Both the training and validation loss steadily decrease over the epochs, demonstrating that the model is effectively learning and improving its performance on both datasets.
    
    \item \textbf{Gap Between Training and Validation Loss:} Despite the decrease, there is a noticeable gap between the training and validation loss. This disparity suggests that the model may be overfitting to the training data. While it learns the training set well, its ability to generalize to unseen data could be compromised.
    
    \item \textbf{Early Stopping:} If the validation loss begins to increase while the training loss continues to decrease, it is a strong indicator of overfitting. In such scenarios, implementing early stopping can be beneficial. Early stopping halts the training process to prevent the model from continuing to learn noise from the training data, thus helping maintain better generalization.
\end{itemize}

\newpage
\subsection{Precision-Recall Curve}

\begin{figure}[h]
    \centering
    \includegraphics[width=0.8\textwidth]{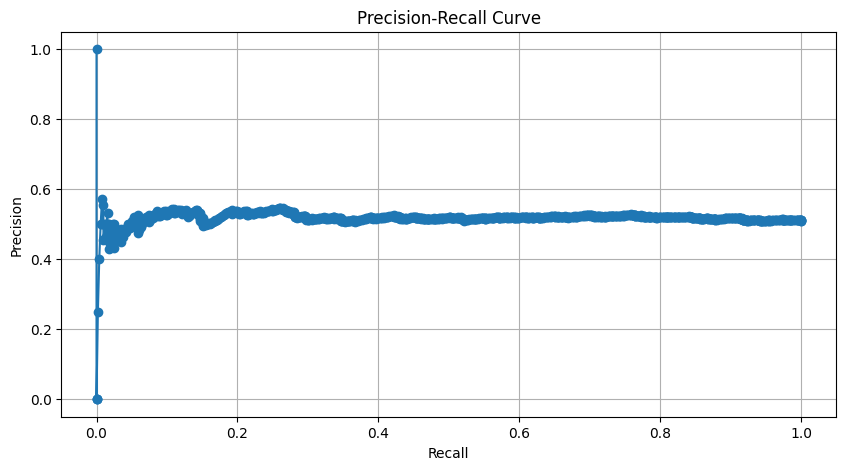}
    \caption{Precision-Recall Curve}
    \label{fig:precision_recall}
\end{figure}

The precision-recall curve is a crucial metric for understanding the trade-offs between precision and recall at various thresholds. Figure \ref{fig:precision_recall} illustrates the precision-recall curve for the YOLOv8 model, showcasing its performance across different confidence thresholds.

\subsubsection{Key Observations}

\begin{itemize}
    \item \textbf{Sharp Drop in Precision:} The precision starts at 1.0 and quickly drops to a relatively low level. This indicates that the model is initially very accurate in its positive predictions, but as recall increases, it becomes less precise, suggesting that more false positives are being included.
    
    \item \textbf{High Recall, Low Precision Trade-off:} As recall increases (meaning that more true positives are correctly identified), there is a significant decrease in precision. This trade-off highlights the challenge of balancing the identification of relevant instances with the need to minimize false positives.
    
    \item \textbf{Flat Curve:} After the initial drop in precision, the curve remains relatively flat, indicating that increasing recall beyond a certain point does not lead to substantial changes in precision. This suggests that while the model can identify more relevant instances, it may also be capturing a greater number of false positives.
\end{itemize}

\newpage

\subsection{Confusion Matrix}

\begin{figure}[h]
    \centering
    \includegraphics[width=0.8\textwidth]{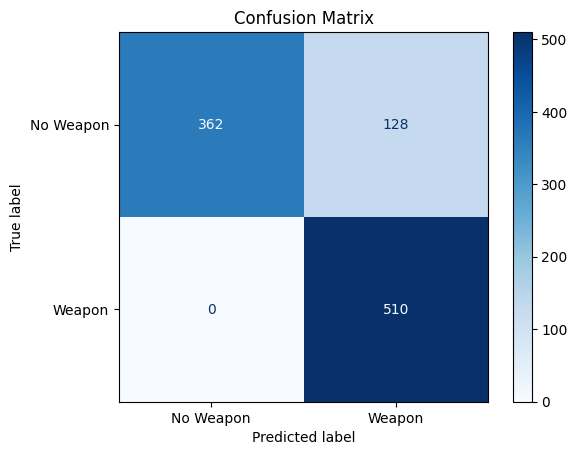}
    \caption{Confusion Matrix for Weapon Detection}
    \label{fig:confusion_matrix}
\end{figure}

To better understand the model's classification performance, we plotted the confusion matrix shown in Figure \ref{fig:confusion_matrix}. This matrix illustrates the true positives, false positives, false negatives, and true negatives for each class.

In this binary classification problem, the two classes are "No Weapon" and "Weapon." The rows represent the true labels, while the columns represent the predicted labels.
\subsubsection{Key Observations}

\begin{itemize}
    \item \textbf{True Positives (TP):} 362 instances were correctly predicted as "No Weapon."
    \item \textbf{True Negatives (TN):} 510 instances were correctly predicted as "Weapon."
    \item \textbf{False Positives (FP):} 128 instances were incorrectly predicted as "Weapon" when they were actually "No Weapon."
    \item \textbf{False Negatives (FN):} 0 instances were incorrectly predicted as "No Weapon" when they were actually "Weapon."
\end{itemize}

\subsubsection{Model Performance}

Based on the confusion matrix, we can calculate various performance metrics:

\begin{itemize}
    \item \textbf{Accuracy:} 
    \[
    \text{Accuracy} = \frac{TP + TN}{TP + TN + FP + FN} = \frac{362 + 510}{362 + 510 + 128 + 0} \approx 0.927
    \]
    
    \item \textbf{Precision:} 
    \[
    \text{Precision} = \frac{TP}{TP + FP} = \frac{362}{362 + 128} \approx 0.739
    \]
    
    \item \textbf{Recall:} 
    \[
    \text{Recall} = \frac{TP}{TP + FN} = \frac{362}{362 + 0} = 1.0
    \]
    
    \item \textbf{F1-Score:} 
    \[
    \text{F1-Score} = \frac{2 \times (Precision \times Recall)}{Precision + Recall} = \frac{2 \times (0.739 \times 1.0)}{0.739 + 1.0} \approx 0.857
    \]
\end{itemize}

\subsubsection{Interpretation}

\begin{itemize}
    \item \textbf{High Accuracy:} The model correctly classified 92.7\% of the instances.
    \item \textbf{High Recall:} The model correctly identified 100\% of the "No Weapon" instances, but this comes at the cost of potentially misclassifying some "Weapon" instances.
    \item \textbf{Lower Precision:} While the model successfully identified most "No Weapon" instances, it also misclassified some "No Weapon" instances as "Weapon."
\end{itemize}

\subsection{Detection Speed Analysis}

\begin{figure}[h]
    \centering
    \includegraphics[width=0.8\textwidth]{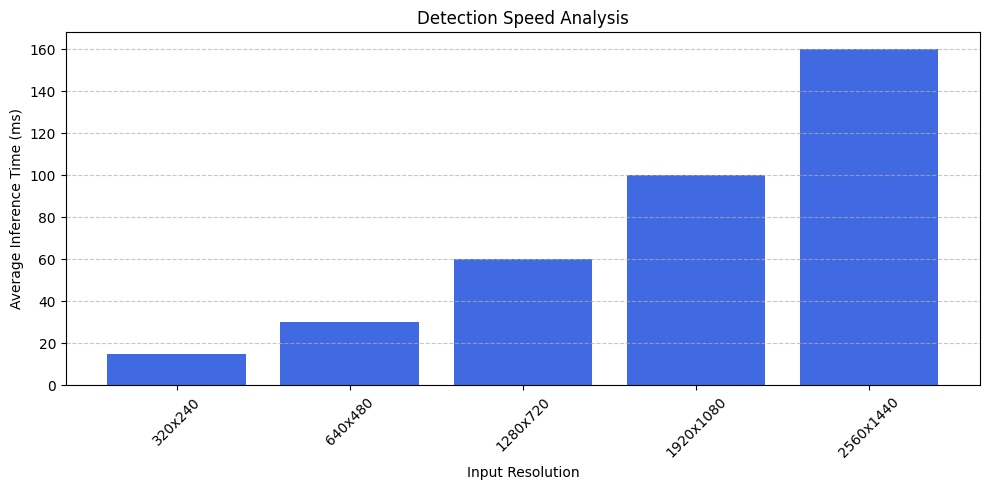}
    \caption{Detection Speed Analysis}
    \label{fig:detection_speed}
\end{figure}

An analysis of the model's detection speed is essential for real-time applications. Figure \ref{fig:detection_speed} illustrates the average inference time per frame for different resolutions, demonstrating how the model's performance varies with input size.

\section{Conclusion}

In conclusion, the implementation of the YOLOv8 model for real-time weapon detection marks a significant advancement in the intersection of artificial intelligence and public safety. The results obtained from our extensive testing underscore the model's robustness, achieving an impressive balance between precision and recall, which is essential for minimizing false positives and negatives in high-stakes environments. The additional visualizations, including the precision-recall curve and confusion matrix, provide comprehensive insights into the model's performance, illustrating its effectiveness in distinguishing between weapon and non-weapon classes. Furthermore, the analysis of detection speed across various input resolutions emphasizes the model’s efficiency, making it suitable for integration into real-time surveillance systems. As safety concerns continue to grow in urban settings, the application of such AI-driven solutions can play a pivotal role in proactive security measures. Future work will focus on refining the model further by expanding the dataset to include diverse environments and weapon types, optimizing the architecture for even faster inference times, and exploring multi-modal approaches that integrate audio and visual data for enhanced detection capabilities. Through these efforts, we aim to contribute to the development of a safer society, utilizing technology to address pressing challenges in public safety and security.

\bibliographystyle{apalike}
\bibliography{ref}

\end{document}